\documentclass[twoside]{article}
\usepackage{aistats2016}
\usepackage[colorlinks = true,urlcolor= blue, linkcolor = red, citecolor= blue]{hyperref}
\usepackage{url}

\usepackage{algorithm}
\usepackage{algorithmic}
\usepackage{amsfonts}
\usepackage{amsmath}
\usepackage[titletoc]{appendix}
\usepackage{bm}
\usepackage{dsfont} 
\usepackage{multicol}
\usepackage{multirow}
\usepackage{booktabs}
\usepackage{arydshln}
\usepackage{gensymb}
\usepackage{wrapfig}
\usepackage{capt-of}
\usepackage{wrapfig}

\usepackage{graphicx} 
\usepackage[update,prepend]{epstopdf}
\usepackage{epsfig}
\usepackage{subfigure} 

\DeclareGraphicsExtensions{.pdf,.eps,.png,.jpg}

\usepackage{natbib}
\begin{document}

%

%

\twocolumn[

\aistatstitle{Generalized Spectral Kernels}

\aistatsauthor{ Yves-Laurent Kom Samo \\ ylks@robots.ox.ac.uk \And Stephen J. Roberts \\ sjrob@robots.ox.ac.uk }

\aistatsaddress{University of Oxford \And University of Oxford} ]

\begin{abstract}
In this paper we propose a family of tractable kernels that is dense in the family of bounded positive semi-definite functions (i.e. can approximate \textit{any} bounded kernel with arbitrary precision). We start by discussing the case of stationary kernels, and propose a family of spectral kernels that extends existing approaches such as \textit{spectral mixture kernels} and \textit{sparse spectrum kernels}. Our extension has two primary advantages. Firstly, unlike existing spectral approaches that yield infinite differentiability, the kernels we introduce allow learning the degree of differentiability of the latent function in Gaussian process (GP) models and functions in the reproducing kernel Hilbert space (RKHS) in other kernel methods.  Secondly, we show that some of the kernels we propose require considerably fewer parameters than existing spectral kernels for the same accuracy, thereby leading to faster and more robust inference. Finally, we generalize our approach and propose a flexible and tractable family of spectral kernels that we prove can approximate \textit{any} continuous bounded nonstationary kernel.
\end{abstract}

\section{Introduction}
Over the past two decades, the use of kernels has been at the heart of many endeavours in the statistics and machine learning communities. Kernels are often used as a flexible way of departing from linear hypotheses in learning machines, thereby allowing for more complex nonlinear patterns (\cite{vapnik95, vapnik98}). They have indeed been successfully applied to problems of classification, clustering, density estimation and regression. The duality between kernels and covariance functions has made kernels a critical tool for both frequentist and Bayesian statisticians.\footnote{We will use the expressions 'kernel' and 'covariance function' interchangeably to denote any symmetric positive semi-definite function. Unless stated otherwise, kernels in this paper are real-valued.} In the Bayesian nonparametric community, kernels are often used as a covariance function of a Gaussian process (GP), introduced as prior over a latent function that is to be inferred from the data. The family of covariance functions postulated for the GP is typically chosen so as to express prior domain knowledge about the underlying function, such as periodicity, regularity and range. The parameters of the kernel are then learned from the data. When one is concerned with automatically uncovering structures from datasets, a flexible family of kernels should be used that can account for intricate patterns. In that regards, it is worth noting that, as most (if not all) loss functions in kernel methods are continuous in the Gram/covariance matrix\footnote{E.g. the negative log-likelihood in GP methods, the Lagrangian in kernel SVM etc...}, if a family of kernels $(k_{\theta_K})_{K \in \mathbb{N}^*}$ can approximate arbitrarily well any continuous bounded kernel, then for any continuous bounded kernel $k$, there exists a kernel $k_{\theta_K^*}$ in the foregoing family that is at least as good as $k$ for the problem at hand (i.e. one that achieves a loss at least as small as that of $k$).

\textbf{Related work}

Approaches have been proposed in recent years that introduce greater flexibility by combining standard unidimensional kernels through series of compounded operations preserving the positive semi-definite property. Examples of such approaches include the \textit{hierarchical kernel learning} model of \cite{hkl}, the \textit{additive kernels} of \cite{DuvenaudNR2012} and the compositional search method of \cite{DuvLloGroetal13}. Although these methods may be major improvements on popular isotropic kernels for some applications, they are limited in that they may not approximate every stationary kernel with arbitrary precision.

The aforementioned limitation has been addressed by spectral approaches such as the \textit{sparse spectrum kernels} of \cite{sparsespectrum} and the \textit{spectral mixture kernels} of \cite{wilson2013gaussian}. Their theoretical underpinning, namely Bochner's theorem (\cite{stein, rasswill, rudin}), is particularly helpful to construct flexible classes of kernels in that it fully characterises all stationary kernels with a relatively simple spectral representation condition.
\begin{theorem}(Bochner's theorem) A complex-valued function $k$ on $\mathbb{R}^d$ is the covariance function of a weakly stationary mean square continuous complex-valued random process on $\mathbb{R}^d$ if and only if it can be represented as 
\begin{equation}
\label{eq:spc_decom}
k(\tau) = \int_{\mathbb{R}^d} e^{2\pi i \omega^T \tau} \mu (d\omega),
\end{equation}
where $\mu$ is a positive finite measure.
\end{theorem}
Bochner's theorem introduces a duality between the flexibility of a class of stationary kernels and the flexibility of the corresponding family of spectral measures $\mu$. The link between  \textit{sparse spectrum kernels} and \textit{spectral mixture kernels} can be understood in the light of Lebesgue's decomposition theorem (\cite{halmos, hewitt}). Lebesgue's decomposition theorem implies that any positive finite measure $\mu$, can be uniquely decomposed as 
\begin{equation}
\label{eq:leb_decomp}
\mu = \mu_{\text{cont.}} +  \mu_{\text{sing.}},
\end{equation}
where $\mu_{\text{cont.}}$ is a finite measure that is absolutely continuous with respect to Lebesgue's measure, and $\mu_{\text{sing.}}$ is a finite measure that is mutually singular with Lebesgue's measure\footnote{That is there exists a partition $\mathbb{R}^d = \mathcal{S}_{\mu_{\text{L}}} \cup \mathcal{S}_{\mu_{\text{sing.}}},  \mathcal{S}_{\mu_{\text{L}}} \cap \mathcal{S}_{\mu_{\text{sing.}}}=\emptyset$ such that every subset of  $\mathcal{S}_{\mu_{\text{sing.}}}$ is of null Lebesgue's measure and for every subset $A \subset \mathcal{S}_{\mu_{\text{L}}}$, $\mu_{\text{sing.}}(A)=0$.}. Examples of positive finite measures that are mutually singular with Lebesgue's measure are the discrete\footnote{Discrete or pure-point measures are measures supported on a countable set.} symmetric measures:
\begin{equation}
\mu_{\text{sing.}} = \sum_{k=1}^{+\infty} \frac{a_k}{2} (\delta_{\omega_k} + \delta_{-\omega_k}), \nonumber
\end{equation}
where $\omega_k \in \mathbb{R}^d, ~a_k \geq 0, ~\sum_{k=1}^{+\infty} a_k < +\infty,$
\begin{align}
\forall A \subset \mathbb{R}^d, \delta_x(A) = \left\{ 
  \begin{array}{l l}
    1 & \text{if } x \in A \\
   0 & \text{if } x \notin A
  \end{array} \right..\nonumber
\end{align}
It follows from Eq. (\ref{eq:spc_decom}) that these measures yield covariance functions of the form:
\begin{equation}
\label{eq:k_sing}
k_{\text{sing.}}(\tau) = \sum_{k=1}^{+\infty} a_k \cos(2\pi \omega_k^T \tau).
\end{equation}
When one is concerned with flexibly learning the shape of the covariance function from the data, the Fourier coefficients $a_k$ need to be inferred directly. For practical purposes we can only work with a finite number $K$ of Fourier coefficients. This gives rise to a simple extension of the \textit{sparse spectrum kernels} introduced by \cite{sparsespectrum}. The authors capped the number of spectral components and required that the Fourier coefficients be identical:
\begin{align}
\label{eq:kss}
k_{\text{SS}}(\tau) = \frac{\sigma^2}{K} \sum_{k=1}^{K} \cos(2\pi \omega_k^T \tau).
\end{align}
However, this family of kernels has three pitfalls. Firstly, they are prone to over-fitting. As an illustration, when used for GP regression, \cite{sparsespectrum} proved that such kernels are equivalent to Bayesian basis function regression with trigonometric basis functions. As such, the learning machine will aim at inferring the $K$ major spectral frequencies evidenced in the training data. This will only lead to appropriate prediction out-of-sample when the underlying latent phenomenon can be appropriately characterized by a finite discrete spectral decomposition that is expected to be the same everywhere on the domain. Secondly, in GP regression, such kernels implicitly postulate that the covariance between the values of the GP at two points does not vanish as the distance between the points becomes arbitrarily large. This imposes \textit{a priori} the view that the underlying function is highly structured, which might be unrealistic in many real-life non-periodic applications. Thirdly, covariance functions of the form of Eq. (\ref{eq:k_sing}) yield infinite differentiability in the mean square sense. As noted by \cite{stein}, this is unrealistic for modelling several physical processes.

Random Fourier features methods (\cite{rahimi07, le13, le15}) are closely related to sparse spectrum kernels. They are based on the observation that Eq. (\ref{eq:spc_decom}) may be rewritten as 
\begin{align}
k(\tau) = \sigma^2 \int_{\mathbb{R}^d} e^{2\pi i \omega^T \tau} \mathbb{P} (d\omega) := \sigma^2 \text{E}_{\mathbb{P}}\left(e^{2\pi i \omega^T \tau}\right), \nonumber
\end{align}
with $\mathbb{P}=\frac{\mu}{\mu \left(\mathbb{R}^d\right)}$ and $\sigma^2 = \mu \left(\mathbb{R}^d\right)$. It then follows that, for any symmetric probability distribution $\mathbb{P}$, if the frequencies $\omega_k$ in Eq. (\ref{eq:kss}) are sampled from $\mathbb{P}$, then the corresponding sparse spectrum kernel $k_{\text{SS}}(\tau)$ (Eq. (\ref{eq:kss})) is an unbiased and consistent estimate of $k(\tau)$. Although random Fourier features methods are scalable, they do not address the need for flexibly learning the spectral measure $\mu$ from the data and are not applicable to nonstationary kernels.

The approach introduced by \cite{wilson2013gaussian} focuses on the continuous part $\mu_{\text{cont.}}$ of Lebesgue's decomposition Eq. (\ref{eq:leb_decomp}). It follows from Radon-Nikodym's theorem (\cite{halmos}) that $\mu_{\text{cont.}}$ admits a (positive) density $f$ with respect to Lebesgue's measure. Moreover, it is easy to see from Eq. (\ref{eq:spc_decom}) that $\int_{\mathbb{R}^d} f(\tau) d\tau = k_{\text{cont.}}(0) < +\infty$. Hence, $\tau \to \frac{f(\tau)}{\int_{\mathbb{R}^d} f(\tau) d\tau}$ is a probability density function, which \cite{wilson2013gaussian} modelled as independent mixtures of Gaussians in each dimension of the spectral domain. The resulting family of \textit{ spectral mixture kernels} reads:
\begin{align}
\label{eq:sm}
k_{\text{SM}}(\tau) = \sum_{k=1}^{K} \sigma_k^2  \exp\left(- 2\pi^2 \vert \vert \tau \odot\gamma_k \vert \vert^2\right) \cos\big(2\pi \omega_k^T \tau \big), 
\end{align}
where $\tau \in \mathbb{R}^d, ~ \omega_k \in \mathbb{R}^{+d},  ~ \gamma_k  \in \mathbb{R}^{+d}, ~  \sigma_k > 0$, and $\tau \odot\gamma_k$ denotes the Hadamard (also known as entrywise) product between the vectors $\tau$ and $\gamma_k$.

Although mixtures of Gaussian distributions can be used to approximate any distribution, spectral mixture kernels are limited in that, when used as covariance functions, they yield infinite differentiability in the mean square sense. Such an excessive smoothness assumption might result in poor predictive accuracy. Moreover, a large number of spectral mixture components might be required to account for lower degrees of smoothness evidenced in the data. This would result in inference techniques that are costlier, and less robust to local optima.

When kernels are used as covariance functions, complex patterns in datasets may also be regarded as evidence of nonstationarity, under the (ergodic) assumption that some properties of a single path, for instance the degree of homogeneity, are the same as the corresponding properties considered across random samples of the underlying process. This approach is common in time series analysis. In Bayesian nonparametrics, nonstationarity may also be introduced to express domain knowledge that vary throughout the input space. However, commonly used approaches such as the input-dependent rescaling of stationary covariance functions of \cite{paciorek2004nonstationary} and spatial deformation of stationary covariance functions (\cite{sampson92, damian01, Schmidt03}), are application specific in that they may not approximate arbitrarily well every covariance function.

The primary contribution of this paper is to propose families of spectral kernels we refer to as \textit{generalized spectral kernels}, that (i) we prove can approximate \textit{any} (possibly nonstationary) bounded kernel, and (ii) allow inference of the degree of differentiability of the corresponding stochastic process when used as a covariance function, or functions in the RKHS in alternative kernel methods such as \textit{support vector machines}. We show that the only (to the best of our knowledge) existing families of kernels that can approximate arbitrarily well any stationary kernel, namely the \text{spectral mixture kernels} of \cite{wilson2013gaussian} and the \text{sparse spectrum kernels} of \cite{sparsespectrum}, are special cases of the families we propose. 

The rest of the paper is structured as follows. In section \ref{sct:gs} we introduce \textit{generalized spectral kernels}, and we prove that they can approximate any continuous bounded kernel. We start by providing the intuition and mathematical background underpinning our approach in section \ref{sct:int}. In section \ref{sct:stat} we introduce stationary \textit{generalized spectral kernels}, we prove that they can approximate arbitrarily well any stationary kernel, we show that they extend existing approaches, and we provide examples of stationary \textit{generalized spectral kernels} that allow the learning of the degree of differentiability of latent functions. In section \ref{sct:nonstat} we extend our approach to nonstationary kernels, and we prove that the family of \textit{generalized spectral kernels} we introduce can approximate arbitrarily well any continuous bounded kernel. We provide empirical evidence that validates our approach in section \ref{sct:exp}, and we conclude with a discussion in section \ref{sct:concl}.
\section{Generalized Spectral Kernels}
\label{sct:gs}
\subsection{Intuition and Background}
\label{sct:int}
The intuition behind our approach is best illustrated with stationary kernels that admit a spectral density. From a practical perspective, in GP models, these are kernels that postulate that the correlation between two GP values vanishes as the distance between the points increases. Considering that spectral measures are finite according to Bochner's theorem, the spectral density of such a kernel is integrable, and hence admits a Fourier transform. In fact, it follows from Bochner's theorem that the spectral density of such a kernel turns out to be its Fourier transform, and vice-versa\footnote{We use the `real' frequency convention for the Fourier transform: $\mathcal{F}(f)(\omega) := \int_{\mathbb{R}^d} e^{-2\pi i \omega^Tx}f(x)d\omega$.}. 

We are interested in constructing families of integrable functions that can `approximate' arbitrarily well any such spectral density in the spectral domain, in a sense that is intuitive and can easily be shown to yield to approximating the inverse Fourier transform (i.e. the kernel in the original domain). This will then allow us to conclude that the inverse Fourier transforms of the approximating functions in the spectral domain can approximate any stationary kernel with absolutely continuous spectral measure. We would also like the family of approximating functions in the original domain to approximate arbitrarily well any stationary kernel whose spectral measure has a non-null singular part in Lebesgue's decomposition (e.g: sparse spectrum kernels). There are two main possible approaches for giving a meaning to the notion of approximating integrable positive-valued functions: one probabilistic and the other deterministic. 

Firstly, noting that integrable positive-valued functions can be normalized to become probability density functions, approximation can be thought of in the sense of the convergence in distribution of random variables. We recall that convergence in distribution is equivalent to pointwise convergence of cumulative density functions, which does not imply convergence of the corresponding probability density functions. Hence, approximating in this sense does not guarantee approximating spectral densities, let alone approximating their inverse Fourier transforms. Stronger notions of convergence of random variables may be used, but the resulting links between approximating in the spectral domain and approximating in the original domain are more involved. This approach is therefore not suitable for our purpose.

The deterministic alternative has several options, two of which are of interest to us. The first notion of approximation is that of the pointwise convergence of functions, according to which a sequence of functions $(f_n)_n$ converges to a function $f$ if and only if for every $x \in \mathbb{R}^d$, the sequence $\left(f_n(x)\right)_n$ converges to $f(x)$. The second notion of approximation is the one of the strong topology of convergence in the space $L^1(\mathbb{R}^d)$ of integrable functions, considered with its canonical norm: \[\forall f,g \in L^1(\mathbb{R}^d), ~ \vert\vert f-g \vert \vert_{L^1} = \int_{\mathbb{R}^d} \vert f(x)-g(x)\vert dx.\]
More precisely, we say of a sequence of integrable functions $(f_n)_n$ that it converges in the $L^1$ sense to an integrable function $f$ if and only if $\vert\vert f-f_n \vert \vert_{L^1}$ converges to 0; in other words when the volume between the surfaces $z=f(x)$ and $z=f_n(x)$ goes to zero. We recall that a set $\mathcal{G}$ is dense in a set  $\mathcal{H}$ with respect to some sense of convergence (topology) if any element $h \in \mathcal{H}$ is the limit of some sequence of elements $g_n$ in $\mathcal{G}$. If $f$ and $f_n$ are integrable, denoting $F$ and $F_n$ their Fourier transforms, it follows from Jensen's inequality that 
\begin{align}
\vert f(x) - f_n(x) \vert &= \vert\int_{\mathbb{R}^d}  e^{2\pi i \omega^Tx}\left(F(\omega)-F_n(\omega)\right)  d\omega \vert \nonumber \\
&\leq \int_{\mathbb{R}^d}  \vert F(\omega)-F_n(\omega) \vert   d\omega \nonumber \\
& \leq \vert\vert F-F_n \vert \vert_{L^1}.\nonumber
\end{align}
Hence, approximating in the spectral domain in the $L^1$ sense implies approximating in the original domain in the pointwise sense. More importantly, if a family of functions $F_n$ is dense in the space of integrable functions (in the spectral domain) with respect to the convergence in $L^1$ , then the corresponding family of inverse Fourier transforms is also dense in the space of integrable kernels (in the original domain) with respect to the pointwise convergence of functions.
\subsection{Stationary Kernels}
\label{sct:stat}
Conditions for a family of functions to be dense in $L^1$ have been extensively studied in the mathematical analysis literature. The most famous results on the matter are known as Wiener's Tauberian theorems (\cite{wiener, rudin, kor}). We recall the theorem that is of interest to us below.
\begin{theorem}
\label{theo:win}
(Wiener's Tauberian theorem) If $f$ is a function in $L^1(\mathbb{R}^d)$, a necessary and sufficient condition for the set of all linear combinations of translations of $f$ to be dense in $L^1(\mathbb{R}^d)$ (in the sense of the convergence  in $L^1$) is that the Fourier transform of $f$ \[F(\omega) := \mathcal{F}(f)(\omega) =  \int_{\mathbb{R}^d}  f(x) e^{-2\pi i \omega^T x} dx\] has no zeros.
\end{theorem}
Gaussian probability density functions in the spectral domain satisfy the conditions of Wiener's Tauberian theorem, and the corresponding linear combinations of translations give rise to the \textit{spectral mixture kernels} of \cite{wilson2013gaussian}.  Wiener's Tauberian theorem however provides a considerably weaker condition. We use it in the spectral domain to construct a broad range of families of tractable functions in the original domain, that are dense in the family of stationary real-valued kernels with respect to the pointwise convergence of functions.
\begin{theorem}
\label{theo:fund}
Let $h$ be a real-valued positive semi-definite, continuous, and integrable function such that $\forall \tau \in \mathbb{R}^d, h(\tau)>0$. The family of functions 
\begin{align}
k_K(\tau) := \sum_{k=1}^{K} \alpha_k  h(\tau \odot\gamma_k ) \cos(2\pi \omega^T_k\tau), 
\end{align}
with $\omega_k, \gamma_k \in \mathbb{R}^{+d}, \alpha_k \in \mathbb{R}, ~ K \in \mathbb{N}^{*}$ is dense in the family of stationary real-valued kernels with respect to the pointwise convergence.
\end{theorem}
\begin{proof}
Sketch: The functions $k_K$ arise as inverse Fourier transforms of linear combinations of translations of the Fourier transform of $h$: $\mathcal{F}(h)$. As $h=\mathcal{F}\left(\mathcal{F}(h)\right)$, the requirement $\forall \tau \in \mathbb{R}^d,~ h(\tau) >  0$ makes Wiener's Tauberian theorem applicable. See \ref{app:proof_fund} for the full proof.
\end{proof}
The assumptions of Th. \ref{theo:fund} are mostly standard. From a practical perspective, the requirement $h(\tau) >  0$ is what makes the family of functions approximate arbitrarily well the absolutely continuous part of any spectral measure, whereas the continuity assumption implies $\underset{\gamma \to 0}{\lim}~ h(\tau \odot \gamma) = h(0) < +\infty$, which allows approximating arbitrarily well the singular part of any spectral measure. The parameters $\gamma_k$ serve as inverse input scales. Noting that $h(\tau \odot \gamma_k) \cos(2\pi \omega^T_k\tau)$ is positive semi-definite, to restrict ourselves to valid kernels we only consider linear combinations with non-negative coefficients. We may also further impose $h(0)=1$ without loss of generality.
\begin{definition} Following the notations of Th. \ref{theo:fund}, we denote \textit{stationary generalized spectral kernels} functions of the form:
\begin{equation}
\label{eq:s_gs}
k(\tau) = \sum_{k=1}^{K} \sigma_k^2  h(\tau \odot \gamma_k) \cos(2\pi \omega^T_k\tau),
\end{equation}
where $h(0)=1$ and $\sigma_k > 0$.
\end{definition}
\textbf{Differentiability}: When stationary \textit{generalized spectral kernels} are used as covariance functions, the following proposition establishes the degree of smoothness they induce.
\begin{proposition}
\label{prop:diff}
A mean zero stationary Gaussian process with stationary generalized spectral covariance function is $p$ times continuously differentiable in the mean square sense if and only if a mean zero stationary Gaussian process with covariance function $h$ is.
\end{proposition}
\begin{proof}
See \ref{app:proof_diff}.
\end{proof}
%
%
\textbf{Examples}: \textit{Sparse spectrum kernels} correspond to the limit case $\gamma_k \to 0$ with equal $\sigma_k$ terms. Moreover, it follows from Eq. (\ref{eq:sm}) that the \textit{spectral mixture kernels} of \cite{wilson2013gaussian} correspond to the special case $h(\tau) = \exp(-2\pi^2 \vert \vert \tau \vert \vert^2)$, which satisfies the conditions of Th. \ref{theo:fund}, and yields infinitely differentiable GPs as a result of Prop. \ref{prop:diff}. It is easy to verify that the Mat\'{e}rn kernels 
\[k_{\text{MA}}(\tau; \nu) = \frac{1}{\Gamma(\nu)2^{\nu-1}} \left(\vert\vert \tau \vert \vert \sqrt{2\nu} \right)^{\nu} K_{\nu}\left(\vert\vert \tau \vert \vert \sqrt{2\nu} \right),\]
where $\Gamma$ is the gamma function and $K_\nu$ is the modified Bessel function of second kind, satisfy the conditions of Th. \ref{theo:fund}. Hence, \textit{Mat\'{e}rn spectral kernels} 
\[ k_{\text{SGS-MA}}(\tau) = \sum_{k=1}^{K} \sigma_k^2  k_{\text{MA}}\left(\tau \odot\gamma_k; \nu\right) \cos(2\pi \omega^T_k\tau),\]
with $\omega_k \in \mathbb{R}^{+d}, ~ K \in \mathbb{N}^{*}$ are also dense in the family of stationary kernels, and allow learning the differentiability of the underlying latent function from the data.

\subsection{Nonstationary Kernels}
\label{sct:nonstat}
Bochner's theorem was the cornerstone of the previous section. The spectral characterisation of stationary kernels it provides turned the problem of approximating stationary kernels into that of approximating measures in the spectral domain. Luckily, it turns out that a more general spectral characterisation exists that includes nonstationary kernels (see  \cite[][\S 26.4]{yaglom} and \cite[][\S 37.4]{loeve} for the univariate case and \cite[][pp. 308]{genton} and \cite[][pp. 149]{kakihara} for a generalization).
\begin{theorem}
\label{theo:ext_boch}A complex-valued bounded continuous function $k$ on $\mathbb{R}^d$ is the covariance function of a mean square continuous complex-valued random process on $\mathbb{R}^d$ if and only if it can be represented as 
\begin{equation}
\label{eq:pc_decom}
k(x, y) = \int_{\mathbb{R}^d \times \mathbb{R}^d} e^{2\pi i (\omega_1^Tx - \omega_2^Ty)} \mu_{F} (d\omega_1, d\omega_2),
\end{equation}
where $\mu_F$ is the Lebesgue-Stieltjes measure associated to some positive semi-definite function $F(\omega_1, \omega_2)$ with bounded-variations.
\end{theorem}
When the spectral measure $\mu_F$ has mass concentrated along the diagonal $\omega_1=\omega_2$, we recover Bochner's theorem. We may once again leverage Wiener's Tauberian theorem to construct families of functions in the spectral domain that are dense in $L^1(\mathbb{R}^{2d})$ with respect to the convergence in $L^1$, so that any spectral density can be approximated arbitrarily well. The argument developed in section \ref{sct:int} may once again be used, in conjunction with Th. \ref{theo:ext_boch} rather than Bochner's theorem, to demonstrate that this would correspond to approximating arbitrarily well any bounded kernel in the original domain in the sense of the pointwise convergence of functions. We obtain the following result.
\begin{theorem}
\label{theo:fund2}Let $(x, y) \to k^*(x, y)$ be a real-valued positive semi-definite, continuous,  and integrable function such that $\forall x, y, ~  k^*(x, y) > 0$. The family 
\begin{align}
&k_K(x, y) := \sum_{k=1}^K \alpha_k k^*(x \odot \gamma_k, y \odot \gamma_k) \Psi_k(x)^T\Psi_k(y) \nonumber
\end{align}
where $\Psi_k(x)=\left( \begin{array}{c}
\cos \left(2\pi x^T\omega_k^1\right) + \cos \left( 2\pi x^T\omega_k^2\right)\\
\sin \left( 2\pi x^T\omega_k^1\right) + \sin \left(2\pi x^T\omega_k^2\right)
\end{array} \right)$, with $\gamma_k \in \mathbb{R}^{+d}, \omega_k^1, \omega_k^2 \in \mathbb{R}^{d}, \alpha_k \in \mathbb{R}, ~ K \in \mathbb{N}^{*}$ is dense in the family of real-valued continuous bounded nonstationary kernels with respect to the pointwise convergence of functions.
\end{theorem}
\begin{proof}
See \ref{app:proof_fund2}.
\end{proof}
The functions $k^*(x \odot \gamma_k, y \odot \gamma_k) \Psi_k(x)^T\Psi_k(y)$ are positive semi-definite like products of such functions,  so that to build a flexible family of expressive nonstationary kernels we may simply require $\alpha_k>0$. We may also impose $k^*(0,0)=1$ without loss of generality.
\begin{definition}
Following the notations of Th. \ref{theo:fund2}, we denote \textit{generalized spectral kernels} functions of the form:
\begin{align}
\label{eq:gen_k}
&k(x, y) =\sum_{k=1}^K \sigma_k^2 k^*(x \odot \gamma_k, y \odot \gamma_k) \Psi_k(x)^T\Psi_k(y)
\end{align}
with $\sigma_k > 0$ and $k^*(0,0)=1$, and relaxing the integrability condition on $k^*$.
\end{definition}
\textbf{Remarks:} We note that when $k^*$ is stationary and $\omega_k^1 = \omega_k^2=\omega$ we recover stationary generalized spectral kernels. The differentiability discussion of the stationary case can be extended. In the general setting, it is sufficient that a generalized spectral covariance function be $2p$ times differentiable for the corresponding mean zero Gaussian process to be mean square $p$ times differentiable (\cite{adlertaylor}). Similarly to the stationary case, the degree of smoothness may be learned or set a priori through the function $k^*$.

\textbf{Examples of integrable $k^*$:} \cite{silvermann} introduced numerous examples of real-valued continuous and integrable covariance functions that are so-called `locally stationary'; i.e. of the form \[k^*(x, y) = k_1\left(x-y\right)k_2\left(\frac{x+y}{2}\right),\] where $k_1$ is a stationary covariance function and $k_2$ is positive-valued. The approaches suggested by the author are flexible enough that $k^*$ can be constructed so that $\forall x,y, ~k^*(x, y) >0$ and the degree of differentiability may be controlled for instance by taking $k_1$ to be a Mat\'{e}rn kernel. Moreover, with this choice of $k^*$, nonstationary random Fourier features approximations may easily be constructed by noting that Eq. (\ref{eq:pc_decom}) may be rewritten in this case as   
\[k(x, y) = \sigma^2\text{E}_{\mathbb{Q}}\left(\Psi_{\omega_1, \omega_2}(x)^T\Psi_{\omega_1, \omega_2}(y) \right), \]
with $\Psi_{\omega_1, \omega_2}(x)=\left( \begin{array}{c}
\cos \left(2\pi x^T\omega_1\right) + \cos \left(2\pi  x^T\omega_2\right)\\
\sin \left(2\pi x^T\omega_1\right) + \sin \left(2\pi  x^T\omega_2\right)
\end{array} \right)$,
 $\sigma^2 = \frac{1}{4} \sum_{k=1}^K \sigma_k^2$ and where $\mathbb{Q}$ is a location-scale mixture of the distribution whose density is the Fourier transform of $k^*$, and with mixing probabilities deduced from Eq. (\ref{eq:gen_k}).

$k^*$ may also be chosen to be of the form
\begin{equation}
\label{eq:sep_gsk}
k^*(x, y) = k_1(x) k_1(y),
\end{equation}
where $k_1$ is any continuous, integrable and positive-valued function. In this case, $\left\{ k_1(x) \Psi_k(x) \right\}_{k=1}^K$ form a basis of the RKHS induced by the kernel $k$. As the dimension of the RKHS is finite, exact inference may be achieved in $\mathcal{O}(nK^2)$ time complexity and with $\mathcal{O}(nK^2)$ memory requirement in most kernel methods, where $n$ is the number of samples. Differentiability may once again be controlled by taking $k_1$ to be a Mat\'{e}rn kernel.

\section{Experiments}
\label{sct:exp}
In this section SE denotes the squared exponential kernel (i.e. $k(\tau)= \sigma^2 e^{-\frac{1}{2} || \tau \odot \gamma ||^2}$), SS denotes the sparse spectrum kernel, MA*2 denotes the Mat\'{e}rn */2 kernel, S-* denotes the stationary generalized spectral kernel with modulating kernel (i.e. $h$ in Eq. (\ref{eq:s_gs})) the kernel *, and NS-* denotes the nonstationary generalized spectral kernel of the form Eq. (\ref{eq:sep_gsk}) where $k_1$ is the kernel *. Moreover, unless stated otherwise all spectral kernels have $K=5$ spectral components.

\begin{figure}[!h]
\includegraphics[width=\linewidth]{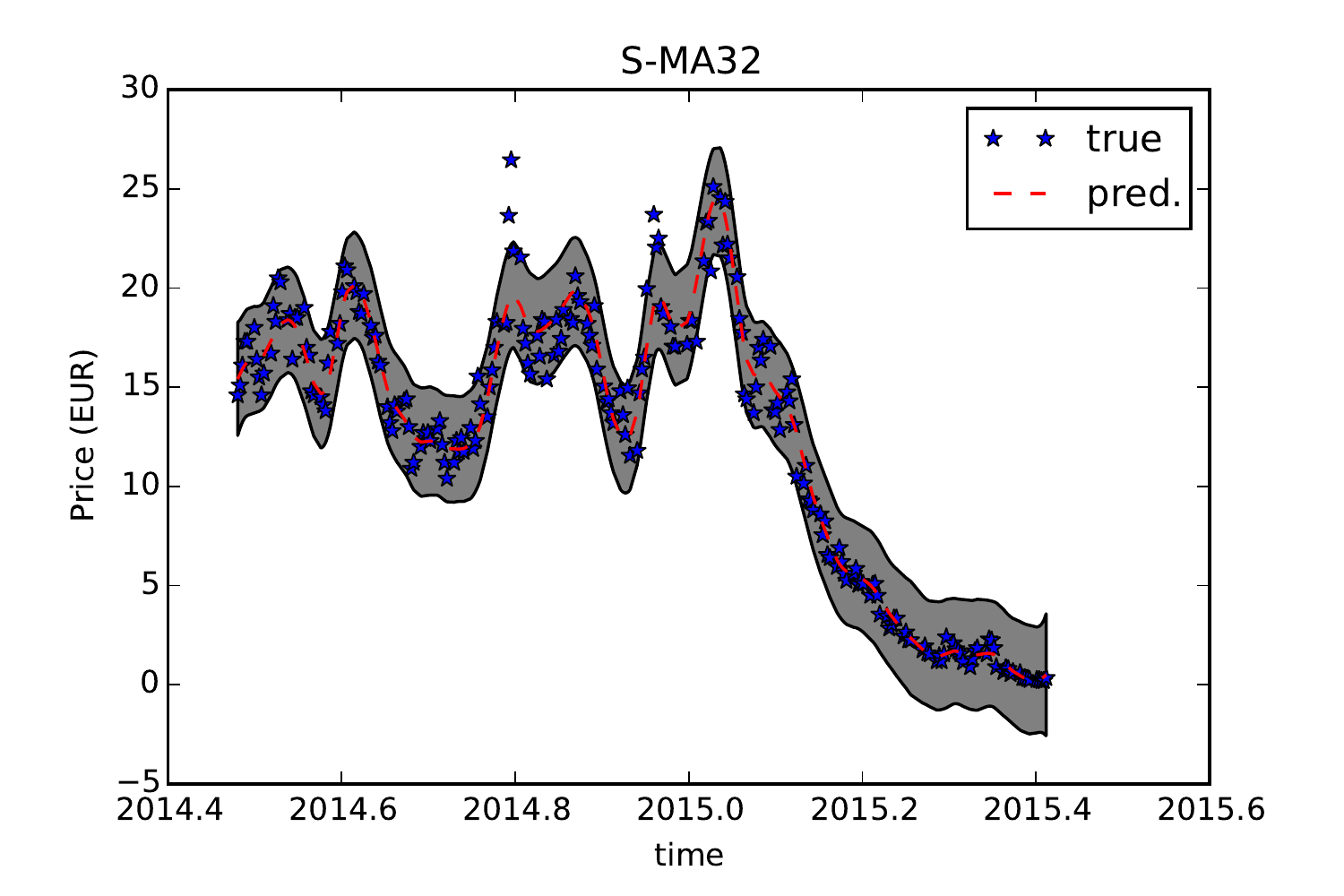}
\caption{Posterior mean $\pm 2\sigma$ in the option experiment with S-MA32 kernel.}
\label{fig:price_opt}
\end{figure}

\begin{figure}[!h]
\includegraphics[width=\linewidth]{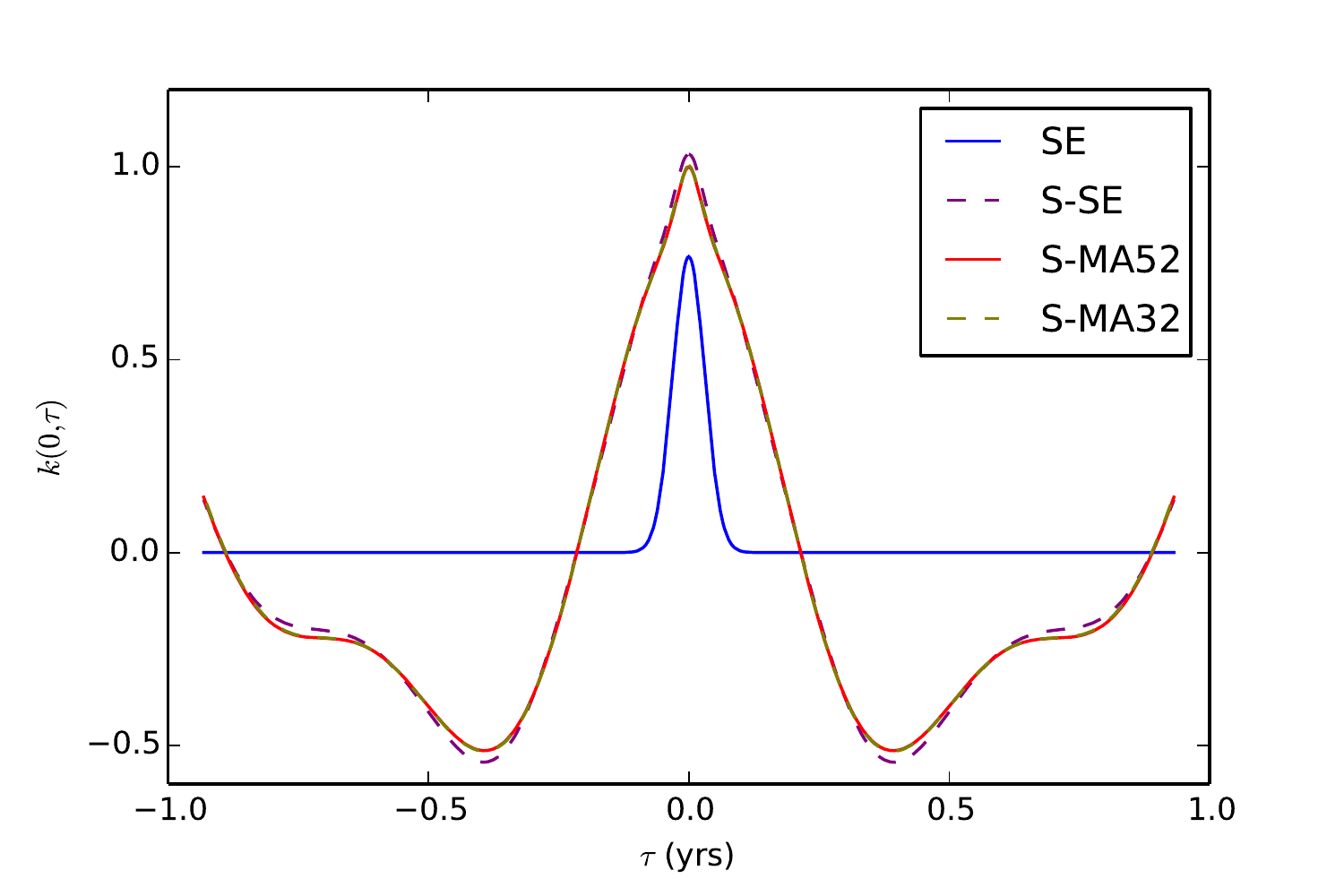}
\caption{Learned kernels in the option experiment.}
\label{fig:kern_opt}
\end{figure}

\textbf{Option pricing}: Firstly, we consider modelling the evolution of the price of a put option on the STOXX Europe 600 Banks index\footnote{We use the strike 195, maturity June 2015 put option on the STOXX 600 Banks index. The data originate from Bloomberg (security code SX7P 6 P195).} as a function of time (i.e. the theta of the option), through GP regression\footnote{Kernel hyper-parameters are learned by maximizing the marginal log-likelihood.}. We use a third of the data for training and the rest for prediction. As evidenced by Tab. \ref{tab:option1}, the \textit{spectral Mat\'{e}rn} 3/2 kernel (S-MA32) improves on the predictive accuracy (RMSE) of the \textit{spectral mixture kernel} (S-SE). Given the density property of S-SE, this suggests that on this dataset, the spectral mixture kernel requires more parameters than the spectral MA 3/2 kernel to achieve the same accuracy, thus making the latter kernel faster and more robust to local maxima during inference. Fig. \ref{fig:price_opt} illustrates the learned posterior mean +/- 2 posterior standard deviations for the S-MA32 kernel. Learned kernels are illustrated in Fig. \ref{fig:kern_opt}.

\begin{table}[h!]
\captionof{table}{Fit and predictive accuracy on the option experiment.}
\label{tab:option1}
\begin{center}
\begin{tabular}{lcccc}
\toprule
	 		& SE 	& S-SE  			& S-MA52 	& S-MA32  \\
\midrule
Log. Lik. 		& -28.56 	& \textbf{-18.61} 	& -19.39 		& -19.60 \\
RMSE 		& 0.89  	& 0.90 			 &  0.76 		& \textbf{0.64}   \\
\bottomrule
\end{tabular}
\end{center}
\end{table}

\textbf{Air temperature anomalies:} Our second experiment is based on the well studied temperature anomalies dataset of \cite{wood_temp}. The dataset consists of monthly readings of air temperature anomalies at various points on the globe in December 1993. The authors defined air temperature anomaly as the deviation of a monthly temperature at a given location from the average over the period 1950-1979 of the monthly temperatures at the same location. There were $445$ readings in December 1993. We selected 2/3 of the data at random for training and predicted the left-out temperature anomalies using GP regression. Training and predictive results are summarized in Tab. \ref{table:air_temp}. We found that the spectral Mat\'{e}rn 1/2 kernel outperforms competing kernels including the spectral mixture kernel (S-SE), which evidences that the latent anomaly function is best modelled as continuous but not smoother. Fig. \ref{table:air_temp} illustrates a map of the posterior mean of the temperature anomaly, and a map of the learned correlation between the temperature anomaly in London and elsewhere on the globe under the spectral Mat\'{e}rn 1/2 kernel. 

\begin{table}[h!]
\captionof{table}{Training log-likelihood and predictive accuracy on the air temperature anomalies dataset.}
\label{table:air_temp}
\begin{center}
\begin{tabular}{lcccc}
\toprule
	 		& SE 		& S-SE 	& S-MA12 		& S-MA32 \\
\midrule
Log. Lik. 		& -358.58 	& -341.68 	& \textbf{-316.89} 	& -326.10 \\
RMSE 			& 1.34 	& 1.31  	&  \textbf{1.28} 	& 1.29 \\
\bottomrule
\end{tabular}
\end{center}
\end{table}
%

\begin{figure}[h]
\includegraphics[width=\linewidth]{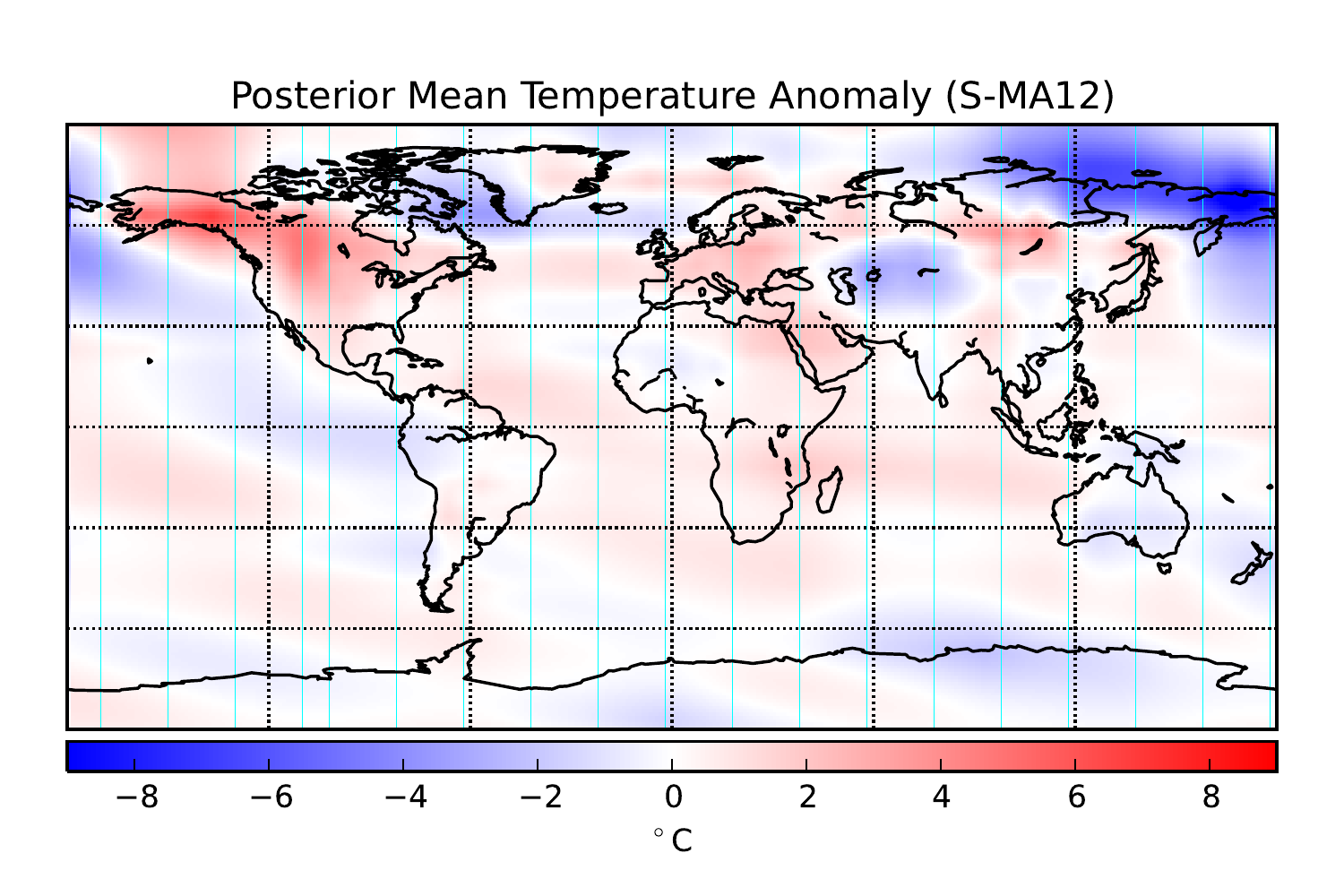}
\includegraphics[width=\linewidth]{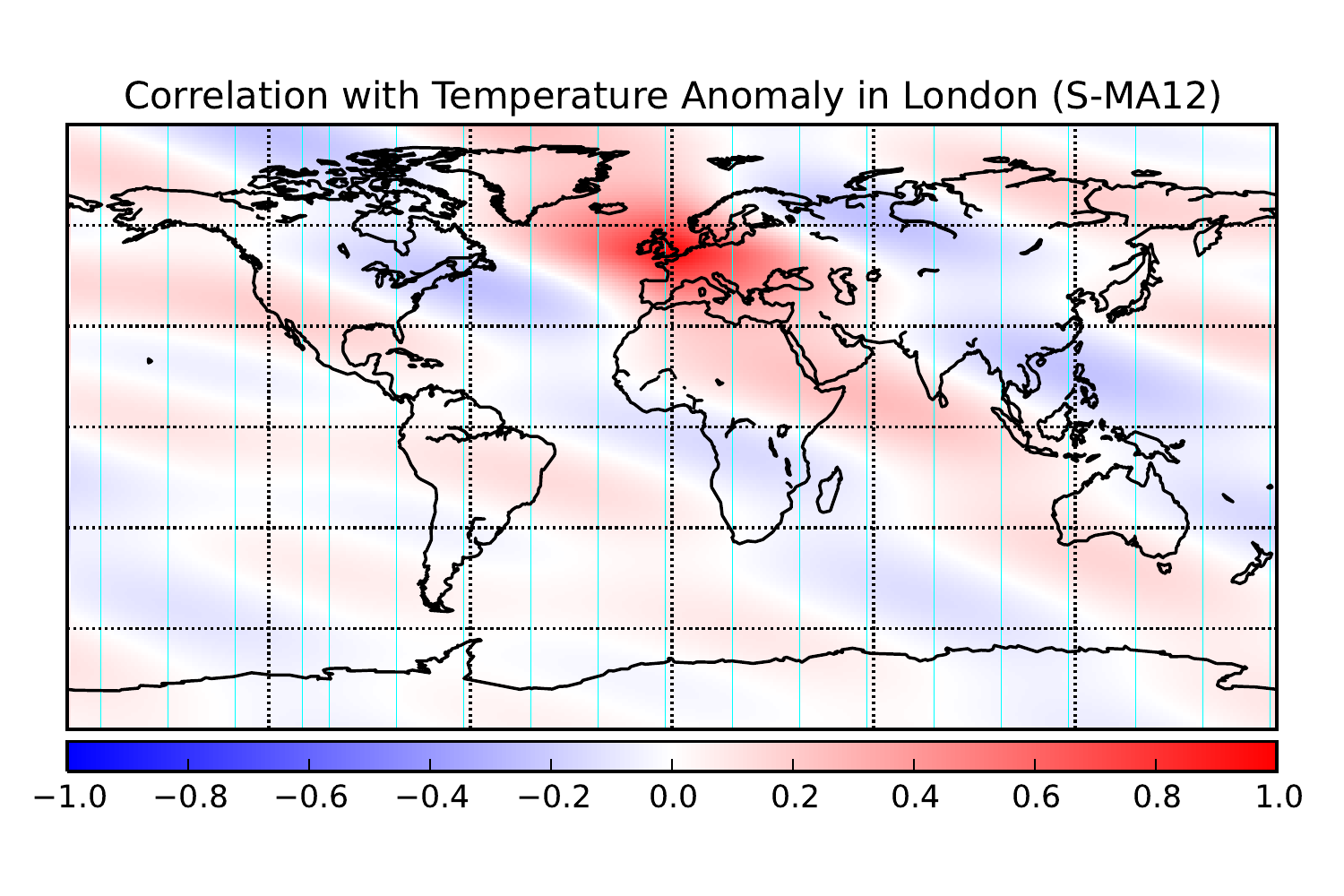} 
\caption{Posterior mean temperature anomaly and learned correlation function between the temperature anomaly in London and elsewhere on the globe under the spectral Mat\'{e}rn 1/2 kernel.}
\label{fig:temp_anom}
\end{figure}

\begin{figure}[!h]
\includegraphics[width=\linewidth]{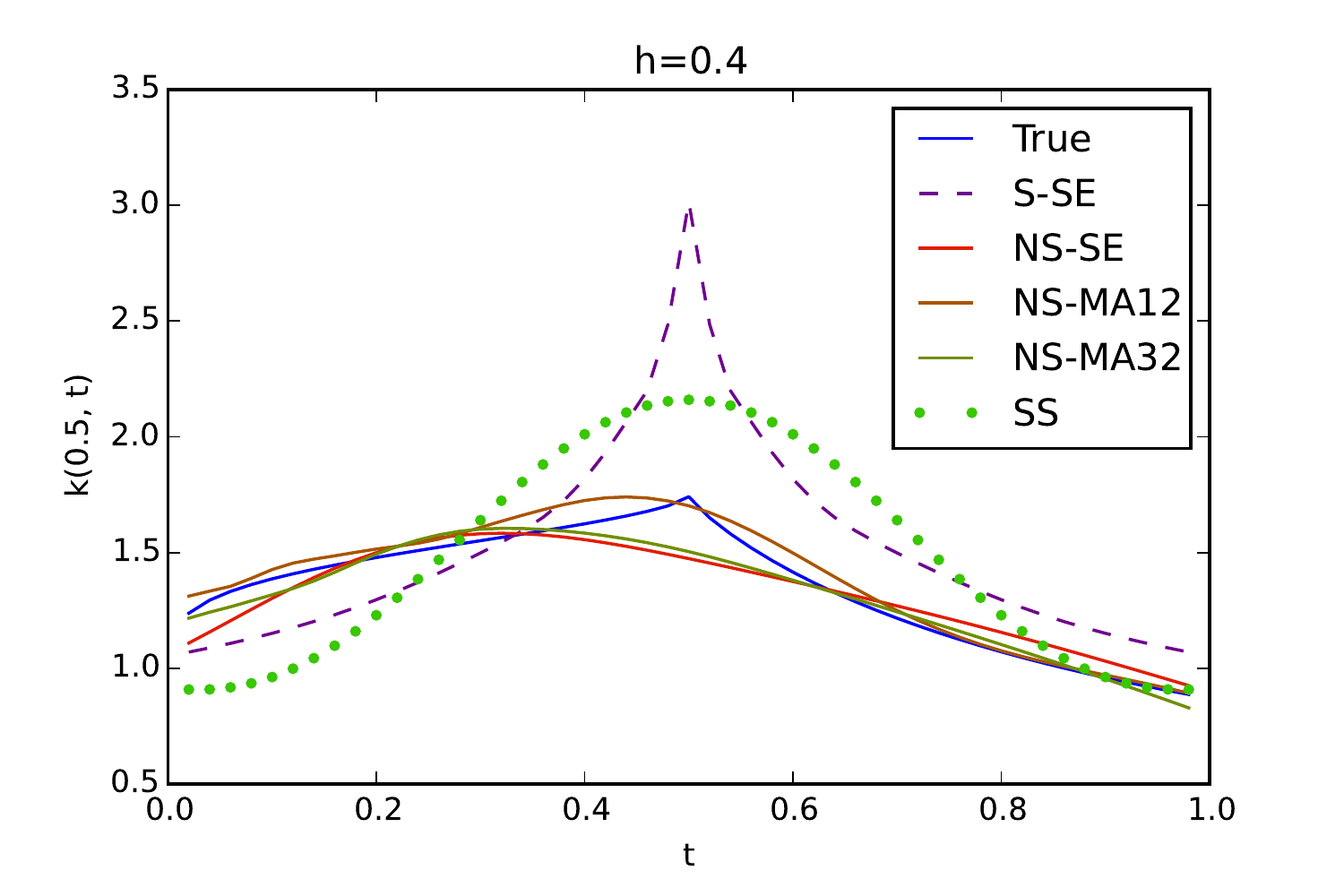}
\includegraphics[width=\linewidth]{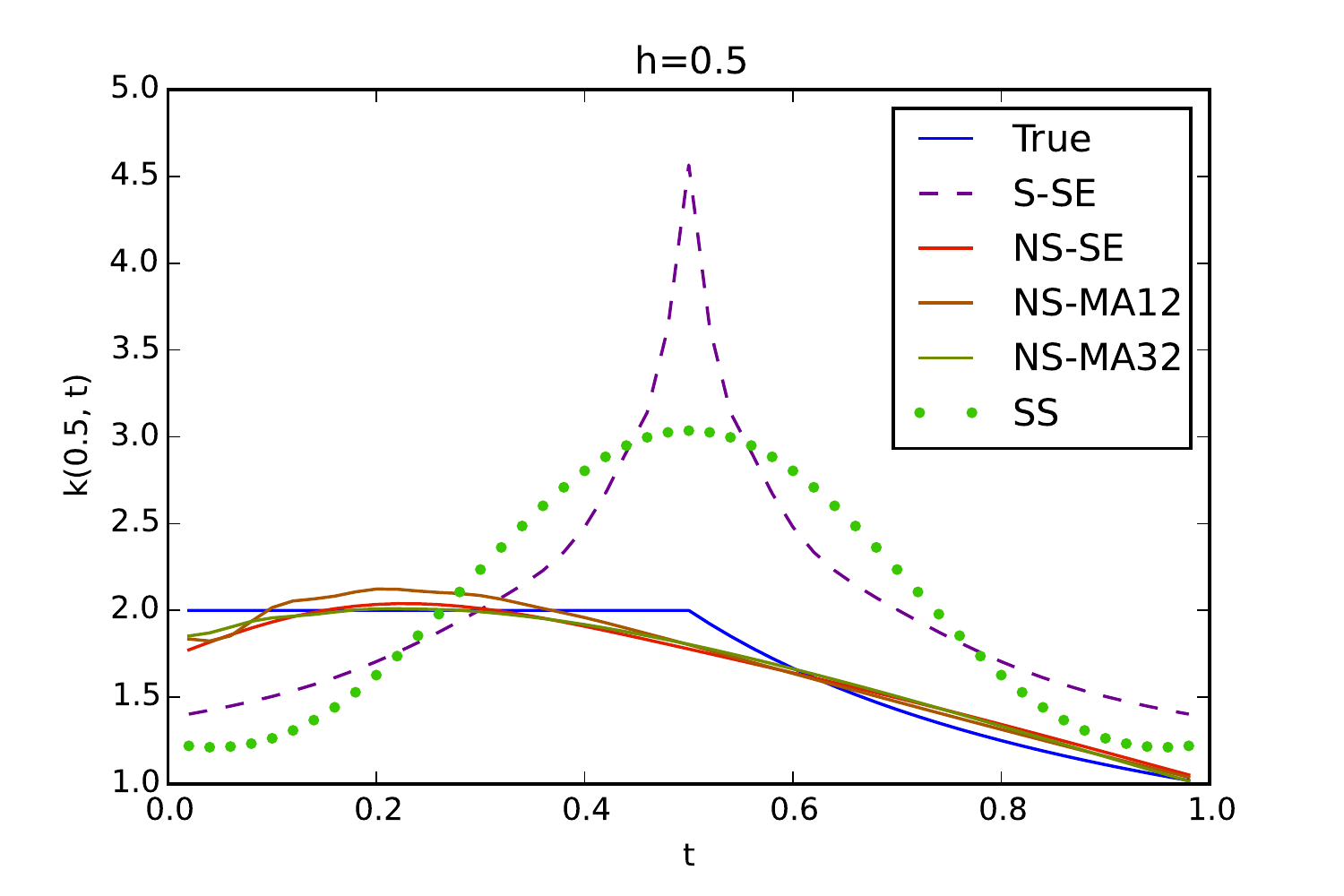} 
\includegraphics[width=\linewidth]{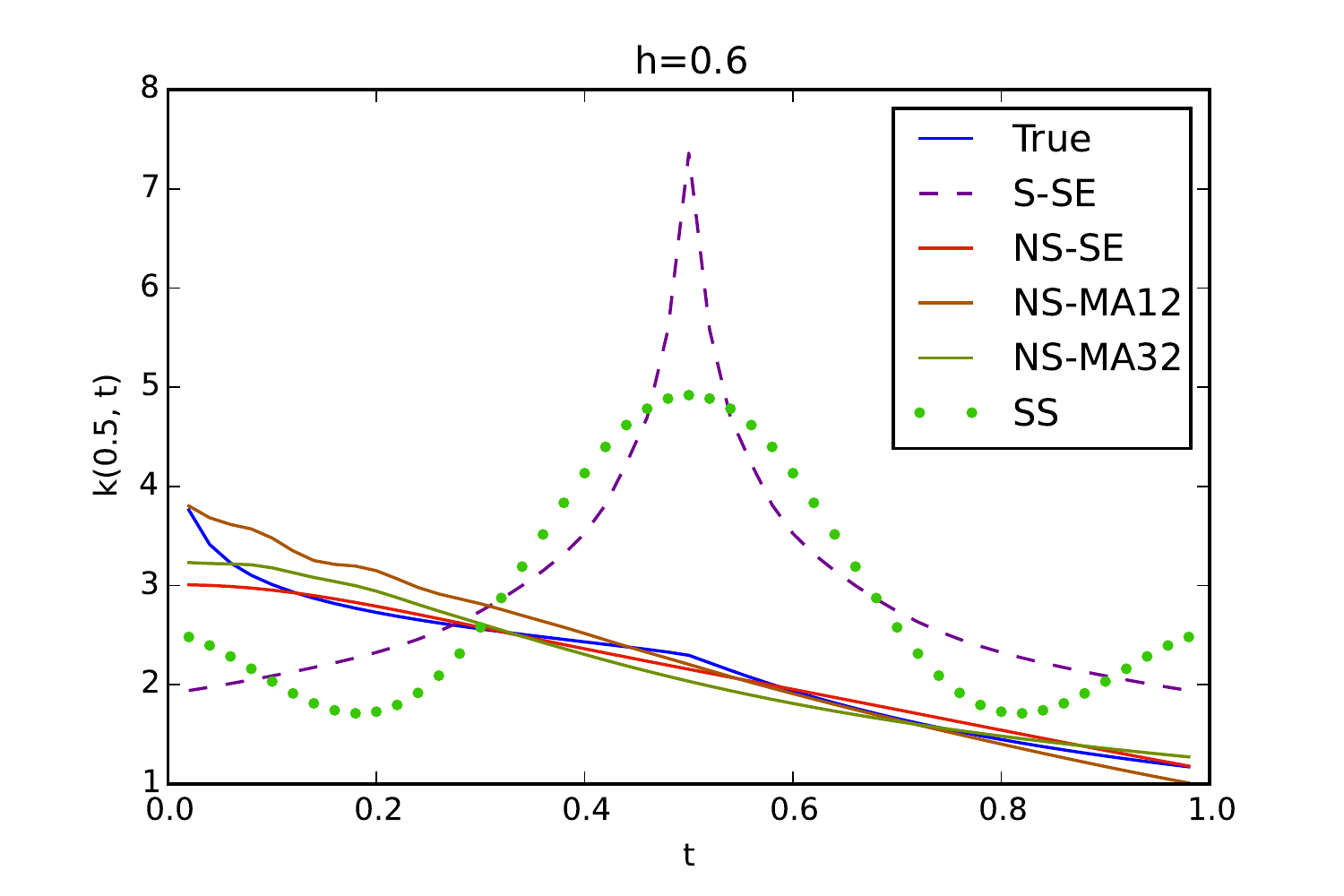}
\caption{Approximations of the IFBM kernel by spectral kernels with $K=5$ components.}
\label{fig:tifbm}
\end{figure}

\textbf{Approximating nonstationary kernels}: Finally, we consider approximating a nonstationary kernel in order to demonstrate the need for nonstationary \textit{generalized spectral kernels}. The nonstationary kernel of interest is the covariance function of a time-inverted fractional Brownian motion (IFBM): 
\[k_{\text{IFBM}}(t, s)= \frac{1}{2}\left(\frac{1}{t^{2h}} +  \frac{1}{s^{2h}} - \left| \frac{1}{t} - \frac{1}{s} \right|^{2h}\right),\]
 with $t, s > 0, ~ 0<h<1$. Such kernels might be particularly useful to model continuous latent functions with known long range behaviour and uncertain short range behaviour (e.g: the price of an option contract as a function of time, value functions in dynamic programming). Higher values of the Hurst index $h$ result in more volatile increments and rougher paths of the corresponding IFBM. We approximate $k_{\text{IFBM}}$ on $(0.01, 1]$ with \textit{generalized spectral kernels} for several values of $h$. The parameters of approximating kernels are learned by minimizing the sum of the square errors between $k_{\text{IFBM}}$ and the approximating spectral kernel, both evaluated on a uniform grid with mesh size $0.02$. Tab. \ref{table:ifbm_approx} illustrates the corresponding root mean square errors normalized by the average value of $k_{\text{IFBM}}$ on the grid for different values of the Hurst index, and with $K=5$ spectral components. Fig. \ref{fig:tifbm} illustrates sections of the learned kernels along the vertical plane $s=0.5$ for $3$ different Hurst indices. It can be seen that nonstationary spectral kernels considerably outperform stationary alternatives such as the \textit{spectral mixture kernel} and the \textit{sparse spectrum kernel}. This comes as no surprise given that $\forall u, ~k_{\text{IFBM}}(0.5, 0.5+u) \neq k_{\text{IFBM}}(0.5, 0.5-u)$, which cannot be modelled by stationary kernels. More importantly, it can be seen at a glance in Fig. \ref{fig:tifbm} that, with only 5 spectral components, nonstationary \textit{generalized spectral kernels} approximate the IFBM kernel pretty well in absolute terms, which is consistent with the density property discussed in the previous section.
\begin{table}[h!]
\captionof{table}{Normalized RMSE of approximations of the time-inverted fractional Brownian motion kernel with Hurst index $h$ by various spectral kernels.}
\label{table:ifbm_approx}
\begin{center}
\begin{tabular}{lccccc}
\toprule
h		& S-SE & SS & NS-SE & NS-MA12 & NS-MA12   \\
\midrule
$0.2$		& 0.22 & 0.26 & 0.09 & \textbf{0.08} & \textbf{0.08}\\
$0.5$	 	& 0.48 & 0.49  &  0.09 & \textbf{0.07} &  0.08\\
$0.8$	 	& 0.37 & 0.37  &  0.10 & 0.09 & \textbf{0.08}\\
\bottomrule
\end{tabular}
\end{center}
\end{table}
\section{Conclusion}
\label{sct:concl}
We propose families of kernels we refer to as \textit{generalized spectral kernels} that we prove can approximate arbitrarily well any continuous bounded kernel. As a result, given that loss functions in kernel methods are often continuous in the kernel/Gram matrix, \textit{generalized spectral kernels} (out of the box) can perform as well as (if not better than) any hand-crafted kernel of practical interest, stationary or not. We show that the only (to the best of our knowledge) families of kernels that have previously been proposed (that can approximate arbitrarily well any stationary kernel) are special cases of \textit{generalized spectral kernels}. Critically, our extension improves on competing approaches in that it allows learning the degree of smoothness of the latent function in Gaussian process models, or that of functions in the RKHS in other kernel methods. More importantly, the families of kernels we propose are the first families of kernels that can approximate arbitrarily well any bounded continuous nonstationary kernel. Finally, our nonstationary extension is amenable to scalable inference either directly, or through a nonstationary extension of random Fourier features approximations.

\clearpage
\bibliography{gsk}

\newpage
\clearpage
\begin{appendices}
 \toptitlebar
\begin{center}{\centering \Large\bfseries Appendix}\end{center}
 \bottomtitlebar
\renewcommand\thesection{Appendix \Alph{section}}

In what follows, we use the `real' (as opposed to angular) frequency convention for the Fourier transform. That is, if $f$ is an integrable real-valued function on $\mathbb{R}^d$ (i.e. $\int_{\mathbb{R}^d} \vert f(x) \vert dx < +\infty$), the Fourier transform of $f$ reads
\[F(\omega) = \mathcal{F}(f)(\omega) =  \int_{\mathbb{R}^d}  f(x) e^{-2\pi i \omega^T x} dx,\]
and the inverse Fourier transform is obtained as 
\[f(x) = \int_{\mathbb{R}^d}  F(\omega) e^{2\pi i \omega^T x}  d\omega.\]
It is a direct consequence of our convention that $\mathcal{F}(F)(x)=f(-x)$.

\section{Proof of Th. \ref{theo:fund}}
\label{app:proof_fund}
We want to prove the following theorem.

\textbf{Theorem}: Let $h$ be a real-valued positive semi-definite, continuous, and integrable function such that $\forall \tau \in \mathbb{R}^d, ~h(\tau)>0$. The family of functions 
\begin{align}
k_K(\tau) := \sum_{k=1}^{K} \alpha_k  h(\tau \odot\gamma_k ) \cos(2\pi \omega^T_k\tau), 
\end{align}
with $\omega_k, \gamma_k \in \mathbb{R}^{+d}, \alpha_k \in \mathbb{R}, ~ K \in \mathbb{N}^{*}$ is dense in the family of stationary real-valued kernels with respect to the pointwise convergence.
\begin{proof} Let us define 
\[h_{\gamma}(\tau) := h(\tau \odot \gamma)\]
and
\[k_{\text{g}}(\tau; \gamma, \omega) := h_{\gamma}(\tau) \cos(2\pi \omega^T\tau).\]
Firstly we note that the function $k_{\text{g}}$ is integrable\footnote{Because the function $h_\gamma$ is and the cosine function is bounded.}, and therefore admits a Fourier transform. As $h$ is integrable and even, it admits an even Fourier transform. Denoting $H = \mathcal{F}(h)$ the Fourier transform of $h$, and $\oslash$ the element-wise division, by definition of the inverse Fourier transform, and using properties of the Fourier transform, it follows that
\begin{align}
\mathcal{F}(k_{\text{g}})(\omega) =  \frac{1}{\prod_{j=1}^d \gamma[j]} \frac{1}{2} &\bigg[H\left((\omega - \omega_k) \oslash \gamma \right)\nonumber \\
&+ H\left((\omega + \omega_k)\oslash \gamma \right)\bigg] \nonumber
\end{align}
which, because $H$ is even, can be rewritten as 
\begin{align}
\mathcal{F}(k_{\text{g}})(\omega) =  \frac{1}{\prod_{j=1}^d \gamma[j]} \frac{1}{2} &\bigg[H\left((\omega_k - \omega) \oslash \gamma \right)\nonumber \\
&+ H\left((\omega_k + \omega)\oslash \gamma \right)\bigg] \nonumber
\end{align}
Let us now consider a stationary real-valued kernel $k$.\\

\textbf{\underline{Case 1}: $\mu_{\text{sing.}} = 0$}

When the singular part of the spectral measure of $k$ in Lebesgue's decomposition theorem is null, the spectral measure $\mu$ of $k$ admits a density with respect to Lebesgue's measure and we note \[\frac{d\mu}{d\omega} = F.\]
The density $F$ is even as $k$ is real-valued. It is also integrable\footnote{$\int_{\mathbb{R}^d}  F(\omega) d\omega = k(0) < +\infty.$}.
Let us consider the function $\omega \to \frac{1}{\prod_{j=1}^d \gamma[j]}H(\omega \oslash \gamma)$. It is integrable and its Fourier transform $\tau \to h_\gamma(-\tau)$ is strictly positive everywhere. Hence, by Wiener's Tauberian theorem, 
\begin{align}
&\exists ~ (\alpha_k, \omega_k)_{k \in \mathbb{N}^{*}}, \alpha_k \in \mathbb{R}, \omega_k \in \mathbb{R}^{d}, \text{s.t.} \nonumber \\
&\forall \omega \in \mathbb{R}^{d}, F(\omega) = \sum_{k=1}^{+\infty} \alpha_k \frac{1}{\prod_{j=1}^d \gamma[j]} H((\omega + \omega_k) \oslash \gamma),\nonumber
\end{align}
where the convergence is to be understood in the $L^1$ sense. As $F$ is even, we also have
\[F(\omega) = F(-\omega) =\sum_{k=1}^{+\infty} \alpha_k \frac{1}{\prod_{j=1}^d \gamma[j]} H\left((-\omega + \omega_k) \oslash \gamma\right),\]
so that 
\begin{align}
F(\omega) = \sum_{k=1}^{+\infty} \alpha_k \frac{1}{2} \frac{1}{\prod_{j=1}^d \gamma[j]}&\bigg[H\left((\omega_k-\omega) \oslash \gamma\right) \nonumber \\
&+ H\left((\omega_k + \omega)  \oslash \gamma \right)\bigg].\nonumber
\end{align}
This proves that the family of spectral density functions 
\[ \sum_{k=1}^{K} \alpha_k \frac{1}{2} \frac{1}{\prod_{j=1}^d \gamma[j]}\bigg[H\left((\omega_k-\omega) \oslash \gamma\right)+ H\left((\omega_k + \omega)  \oslash \gamma \right)\bigg]\] 
is dense in the family of Fourier transforms of integrable stationary kernels\footnote{We recall that Bochner's theorem implies that the spectral density function of an integrable kernel is its Fourier transform.} with respect to the convergence in $L^1$ of functions. Hence, as proved in the paper, the corresponding family of inverse Fourier transforms, namely
\[\sum_{k=1}^{K} \alpha_k  h_{\gamma}(\tau) \cos(2\pi \omega^T_k\tau),\]
is dense in the family of integrable real-valued stationary kernels with respect to the pointwise convergence of functions. This result also holds for the superset $\sum_{k=1}^{K} \alpha_k  h_{\gamma_k}(\tau) \cos(2\pi \omega^T_k\tau)$ by definition of the density property. Moreover, as the cosine function is even, the density property is preserved after imposing the constraint $\omega_k \in \mathbb{R}^{+d}$. \\

\textbf{\underline{Case 2}: $\mu_{\text{cont.}} = 0$}

We now deal with the case where the continuous part of the spectral measure of $k$ in Lebesgue's decomposition theorem is null. A refined version of Lebesgue's decomposition theorem states that the singular measure can be uniquely decomposed as $\mu_{\text{sing.}} = \mu_{\text{pp.}} + \mu_{\text{sc.}}$ where $\mu_{\text{pp.}}$ is a discrete (pure-point) measure and $\mu_{\text{sc.}}$ is mutually singular with Lebesgue's measure. The singular continuous measure $\mu_{\text{sc.}}$ is not intuitive as it gives null probability mass to any countable set of `outcomes', and yet it gives positive probability mass to some sets of outcomes with null `volume' (Lesbegue's measure). For those reasons, we believe singular continuous measures to be of limited interest in most statistical inference problems involving stationary kernels, and we will restrict our attention to discrete measures in this section (i.e. $\mu_{\text{sc.}}=0$).

We recall from Eq. (\ref{eq:k_sing}) that the stationary covariance functions arising from discrete positive and symmetric spectral measures can be written as:
\[
k_{\text{sing.}}(\tau) = \sum_{k=1}^{+\infty} \alpha_k \cos(2\pi \omega_k^T \tau),
\]
with $\alpha_k \geq 0$ and $\sum_{k=1}^{+\infty} \alpha_k < +\infty$. Moreover, as $h$ is positive-valued and positive semi-definite, we have that 
\[\text{det} \left( \begin{bmatrix} 
h(0) & h(\tau \odot \gamma)\\
h(\tau \odot \gamma)  & h(0)
\end{bmatrix} \right) \geq 0,\]
which implies \[\forall \gamma > 0, ~ 0 <h_\gamma(\tau) \leq h(0).\]
Hence, \[\forall \gamma, ~  \vert \frac{\alpha_k}{h(0)} k_{\text{g}}(\tau; \gamma, \omega_k) \vert \leq \vert  \alpha_k \cos(2\pi \omega_k^T \tau) \vert.\]
As $ \sum_{k=1}^{+\infty} \vert \alpha_k \cos(2\pi \omega_k^T \tau) \vert  \leq \sum_{k=1}^{+\infty} \alpha_k < +\infty$, by the dominated convergence theorem we have that
\begin{align}
 \underset{\gamma \to 0}{\lim}~ \sum_{k=1}^{+\infty}  \frac{\alpha_k}{h(0)} k_{\text{g}}(\tau; \gamma, \omega_k) &=  \sum_{k=1}^{+\infty} \underset{\gamma \to 0}{\lim}~   \frac{\alpha_k}{h(0)} k_{\text{g}}(\tau; \gamma, \omega_k) \nonumber \\
 &=  k_{\text{sing.}}(\tau).\nonumber
\end{align}
Hence, any stationary kernel whose spectral measure is a pure-point measure is the limit of kernels of the form $\sum_{k=1}^{K} \alpha_k  h_{\gamma_k}(\tau) \cos(2\pi \omega^T_k\tau)$ (as $\gamma_k$ go to $0$ and $K$ goes to $+\infty$), which concludes the proof in the second case.\\

\textbf{\underline{Case 3}: $\mu_{\text{sing.}} \neq 0$ and $\mu_{\text{cont.}} \neq 0$}

In the general case, we decompose any covariance function $k$ as \[k = k_{\text{sing.}} + k_{\text{cont.}},\] with $k_{\text{sing.}}(\tau) = \int_{\mathbb{R}^d} e^{2\pi i \omega^T \tau} d\mu_{\text{sing.}} (\omega)$,  and $ k_{\text{cont.}}(\tau) = \int_{\mathbb{R}^d} e^{2\pi i \omega^T \tau} d\mu_{\text{cont.}} (\omega)$. 
We then use the two cases previously discussed to conclude that $k$ is the limit of linear combinations of kernels of the form $k_{\text{g}}(\tau; \gamma, \omega)$.
\end{proof}
\section{Proof of Prop. \ref{prop:diff}}
\label{app:proof_diff}
We now prove the following proposition.

\textbf{Proposition}: A mean zero stationary Gaussian process with stationary generalized spectral covariance function is $p$ times continuously differentiable in the mean square sense if and only if a mean zero stationary Gaussian process with covariance function $h$ is.
\begin{proof} $p$ times differentiability of a stationary GP in the mean square sense is equivalent to $2p$ times differentiability of its covariance function at $0$ (\cite{adlertaylor}). It is easy to see that if $h$ is $2p$ times differentiable at 0, then so will the corresponding stationary generalized spectral kernel. Reciprocally, a simple reasoning by contradiction allows us to conclude that if $h$ is not at least $2p$ times differentiable at $0$, $h_{\gamma_k}(\tau) \cos(2\pi \omega^T_k\tau)$ and subsequently the corresponding stationary spectral kernel cannot be.
\end{proof}

\section{Proof of Th. \ref{theo:fund2}}
\label{app:proof_fund2}
In this section we prove Th. \ref{theo:fund2}, which we recall below.

\textbf{Theorem} Let $(x, y) \to k^*(x, y)$ be a real-valued positive semi-definite, continuous,  and integrable function such that $\forall x, y, ~  k^*(x, y) > 0$. The family 
\begin{align}
&k_K(x, y) := \sum_{k=1}^K \alpha_k k^*(x \odot \gamma_k, y \odot \gamma_k) \Psi_k(x)^T\Psi_k(y) \nonumber
\end{align}
where $\Psi_k(x)=\left( \begin{array}{c}
\cos \left(2\pi x^T\omega_k^1\right) + \cos \left(2\pi  x^T\omega_k^2\right)\\
\sin \left( 2\pi x^T\omega_k^1\right) + \sin \left(2\pi  x^T\omega_k^2\right)
\end{array} \right)$, with $\gamma_k \in \mathbb{R}^{+d}, \omega_k^1, \omega_k^2 \in \mathbb{R}^{d}, \alpha_k \in \mathbb{R}, ~ K \in \mathbb{N}^{*}$ is dense in the family of real-valued continuous bounded nonstationary kernels with respect to the pointwise convergence of functions.

\begin{proof}
$k^*$ being integrable, it admits a Fourier transform
\[K^*(\omega_1, \omega_2) := \mathcal{F}(k^*)(\omega_1, \omega_2),\] 
and we have 
\[\forall x,y, ~\mathcal{F}(K^*)(x, y) = k^*(-x, -y) >0.\]
Hence, the conditions of Wiener's Tauberian theorem are met, so that any integrable function on $\mathbb{R}^d \times \mathbb{R}^d$ is the limit of linear combinations of translations of $K^*$.

Let $k$ be a real-valued continuous bounded nonstationary kernel and $\mu_F$ its Lebesgue-Stieltjes spectral measure. We will start with the case where $\mu_F$ is absolutely continuous with respect to Lebesgue's measure. In that case, denoting $f$ the corresponding Radon-Nikodym derivative, we have:
\[k(x, y) = \int e^{2\pi i (x^T \omega_1 - y^T \omega_2)} f(\omega_1, \omega_2) d\omega_1 d\omega_2.\]
Noting that $f$ is integrable, we can define
\[f_K(\omega_1, \omega_2) := \sum_{k=1}^K \beta_k K^*(\omega_1 + \omega_k^1, \omega_2 + \omega_k^2),\]
a sequence of linear combinations of translations of $K^*$ converging to $f$ in the $L^1$ sense. We can always consider such a sequence $\{ f_K \}$  with symmetric functions. In effect, for any candidate $\{f_K\}$, $\{\bar{f}_K\}$ with 
\[\bar{f}_K(\omega_1, \omega_2) = \frac{1}{2} \left(f_K(\omega_1, \omega_2) + f_K(\omega_2, \omega_1)\right),\]
are symmetric, integrable, linear combinations of translations of $K^*$ as $K^*$ is symmetric, and converge to $f$ in the $L^1$ sense. As both $f_K$ and $K^*$ are symmetric, we also have:
\begin{align}
f_K(\omega_1, \omega_2) &= f_K(\omega_2, \omega_1) \nonumber \\
&:= \sum_{k=1}^K \beta_k K^*(\omega_2 + \omega_k^1, \omega_1 + \omega_k^2) \nonumber \\
&= \sum_{k=1}^K \beta_k K^*(\omega_1 + \omega_k^2, \omega_2 + \omega_k^1).\nonumber
\end{align}
We can therefore rewrite $f_K$ as
\begin{flalign}
&f_K(\omega_1, \omega_2) := \nonumber \\
&\sum_{k=1}^K \frac{\beta_k}{2} \bigg(K^*(\omega_1 + \omega_k^1, \omega_2 + \omega_k^2)+ K^*(\omega_1 + \omega_k^2, \omega_2 + \omega_k^1) \nonumber \\
& + K^*(\omega_1 + \omega_k^1, \omega_2 + \omega_k^1) + K^*(\omega_1 + \omega_k^2, \omega_2 + \omega_k^2)\bigg) \nonumber \\
& -\frac{\beta_k}{2} \bigg(K^*(\omega_1 + \omega_k^1, \omega_2 + \omega_k^1) + K^*(\omega_1 + \omega_k^2, \omega_2 + \omega_k^2)\bigg)\nonumber.
\end{flalign}
Denoting 
\[\hat{k}(x, y) :=  \int e^{2\pi i (x^T \omega_1 - y^T \omega_2)} f_K(\omega_1, \omega_2) d\omega_1 d\omega_2,\]
it is  easy to see, by applying Jensen's inequality to $|\hat{k}(x, y)-k(x, y)|$ like we did in the stationary case, that $\hat{k}$ converges to $k$ pointwise. Thus, as $k$ is real-valued, the real-part of $\hat{k}$ converges to $k$ pointwise too. Simple changes of variables using the expanded expression of $f_K(\omega_1, \omega_2)$ give us the following expression for the real-part of $\hat{k}$:
\begin{flalign}
\label{eq:re}
&Re\left(\hat{k}(x, y)\right) = \Bigg[ \sum_{k=1}^K \frac{\beta_k}{2} k^{*}(x, y)\bigg(  \\
& \cos \left( 2\pi x^T (\omega_k^1 - y^T \omega_k^2) \right) + \cos \left(2\pi  (x^T \omega_k^2 - y^T \omega_k^1) \right) \nonumber \\
& + \cos \left( 2\pi (x-y)^T \omega_k^1 \right) + \cos \left( 2\pi (x-y)^T \omega_k^2 \right)\bigg) \Bigg] \nonumber \\
& +\Bigg[\sum_{k=1}^K \frac{-\beta_k}{2} k^{*}(x, y)\cos \left( 2\pi (x-y)^T \omega_k^1 \right) \Bigg] \nonumber \\
&  +\Bigg[\sum_{k=1}^K \frac{-\beta_k}{2} k^{*}(x, y)\cos \left( 2\pi (x-y)^T \omega_k^2 \right) \Bigg] \nonumber
\end{flalign}
If we denote $\Psi_{a,b}(x)=\left( \begin{array}{c}
\cos \left(2\pi x^Ta\right) + \cos \left( 2\pi x^Tb\right)\\
\sin \left( 2\pi x^Ta\right) + \sin \left( 2\pi x^Tb\right)
\end{array} \right)$, by expanding the cosine functions in Eq. (\ref{eq:re}), it follows that each of the three sums above can be rewritten in the form
\[\sum_{k=1}^K \alpha_k k^{*}(x, y) \Psi_{k}(x)^T \Psi_{k}(y),\]
where for the first sum we have $\alpha_k = \frac{\beta_k}{2}$, $\Psi_{k}(x) = \Psi_{\omega_k^1,\omega_k^2}(x)$, for the other two sums $\alpha_k = -\frac{\beta_k}{8}$, for the second sum $\Psi_{k}(x) = \Psi_{\omega_k^1,\omega_k^1}(x)$ and for the last sum $\Psi_{k}(x) = \Psi_{\omega_k^2,\omega_k^2}(x)$. This proves that for any real-valued continuous bounded nonstationary kernel $k$ with absolutely continuous spectral measure, there exist a sequence of the form 
\[k_K(x, y) = \sum_{k=1}^K \alpha_k k^{*}(x, y) \Psi_{k}(x)^T \Psi_{k}(y),\]
that converges to $k$.

As for pure-point spectral measures, they can be written as
\begin{equation}
\label{eq:c1}
\mu_F(A_1, A_2) =  \sum_{k=1}^{+\infty} \beta_k \delta_{\omega_k^1}(A_1) \delta_{\omega_k^2}(A_2),
\end{equation} 
with $\sum_{k=1}^{+\infty} \vert \beta_k \vert < +\infty$. By symmetry of $\mu_F$, Eq. (\ref{eq:c1}) may be rewritten as 
\begin{equation}
\mu_F(A_1, A_2) =  \sum_{k=1}^{+\infty} \frac{\beta_k}{2} \left( \delta_{\omega_k^1}(A_1) \delta_{\omega_k^2}(A_2) + \delta_{\omega_k^2}(A_1) \delta_{\omega_k^1}(A_2)  \right) \nonumber 
\end{equation}
and using the same trick as in the integrable case, we get that the corresponding kernels are of the form
\begin{equation}
k(x, y) := \sum_{k=1}^{+\infty} \alpha_k  \Psi_{k}(x)^T \Psi_{k}(y),\nonumber
\end{equation}
where we also have $\sum_{k=1}^{+\infty} \vert \alpha_k \vert < +\infty$. As both $(x, y) \to \Psi_{k}(x)^T \Psi_{k}(y)$ and $(x, y) \to k^{*}(x \odot \gamma, y \odot \gamma) $  are bounded,\footnote{$k^*$ being continuous and integrable is bounded.} we may once again use the dominated convergence theorem to conclude that 
\begin{align}
&k(x, y) := \sum_{k=1}^{+\infty}   \underset{\gamma \to 0}{\lim}~  \alpha_k k^{*}(x \odot \gamma, y \odot \gamma) \Psi_{k}(x)^T \Psi_{k}(y) \nonumber\\
&=  \underset{\gamma \to 0}{\lim}~\sum_{k=1}^{+\infty}    \alpha_k k^{*}(x \odot \gamma, y \odot \gamma) \Psi_{k}(x)^T \Psi_{k}(y) \nonumber\\
&=  \underset{\gamma \to 0, K \to +\infty}{\lim}~\sum_{k=1}^{K}   \alpha_k  k^{*}(x \odot \gamma, y \odot \gamma) \Psi_{k}(x)^T \Psi_{k}(y) \nonumber,
\end{align}
where we have used the continuity of $k^*$ at $(0, 0)$ and $k^*(0, 0)=1$. This concludes the proof for the pure-point case. Hybrid cases are dealt with in a similar manner to the proof of Th. \ref{theo:fund}.
\end{proof}

\section{Technical discussion on Th. \ref{theo:ext_boch}}
Every real-valued continuous bounded function of the form of Eq. (\ref{eq:pc_decom}) is indeed the covariance function of a real-value mean square continuous stochastic process. Strictly speaking, for a real-valued continuous bounded function that is also the covariance function of a real-valued mean square continuous stochastic process to be of the form of Eq. (\ref{eq:pc_decom}), it has to be \textit{harmonizable} (\cite{yaglom, kakihara}). However, as noted by \cite[][pp. 464]{yaglom}, the only continuous bounded covariance functions known to the author that are not \textit{harmonizable} `are rather complicated and have some unusual, even pathological properties'. Moreover, \cite{kakihara} proved that, if a real-valued bounded function $k$ is the covariance function of a mean square continuous stochastic process $(f(x))$, providing that the mapping $x \to f(x)$ is (strongly) measurable, $k$ is \textit{harmonizable}. The foregoing condition will often be verified by stochastic processes of practical interest, which is the reason why, as did \cite{genton}, we have ignored this technicality in Th. \ref{theo:ext_boch}.

\end{appendices}

\end{document}